\documentclass[10pt,twocolumn,letterpaper]{article}
\usepackage{authblk}
\usepackage{cvpr,times}

\usepackage[utf8]{inputenc} % allow utf-8 inputhttps://v2.overleaf.com/5654625843pmfxvmqsgcxg
\usepackage[T1]{fontenc}    % use 8-bit T1 fonts
\usepackage[breaklinks=true,bookmarks=false]{hyperref}
\usepackage{url}            % simple URL typesetting
\usepackage{booktabs}       % professional-quality tables
\usepackage{nicefrac}       % compact symbols for 1/2, etc.
\usepackage{microtype}      % microtypography
\usepackage{color}
\usepackage[ruled,vlined,algo2e, linesnumbered]{algorithm2e}
\usepackage{wrapfig}
\usepackage{subcaption}
\usepackage{enumitem}
\usepackage{amsfonts,amsthm,amsmath,amssymb,mathtools}

\def\proj{\text{P}}
\def\1{{\bf{1}}}
\def\0{{\bf{0}}}

\def\x{{\bf x}}

\def\cV{\mathcal{V}}
\def\cE{{\mathcal{E}}}
\def\cW{{\mathcal{W}}}
\def\cL{{\mathcal{L}}}

\def\cA{{\mathcal{A}}}
\def\cB{{\mathcal{B}}}
\def\cX{{\mathcal{X}}}
\def\cY{\mathcal{Y}}

\DeclareMathOperator*{\argmin}{arg\,min}

\usepackage{pifont}% http://ctan.org/pkg/pifont
\newcommand{\cmark}{\ding{51}}%
\newcommand{\xmark}{\ding{55}}%

\let\citet\undefined
\newcommand{\citet}{\cite}
\let\citep\undefined
\newcommand{\citep}{\cite}

\renewcommand{\paragraph}[1]{\textbf{#1}~~}

\newcommand\figwidthly{0.1308308428}
\newcommand\figwidthlo{0.1176329946}
\newcommand\figwidtha{0.1778840406}

\usepackage[figuresright]{rotating}
\usepackage[ruled,vlined,algo2e]{algorithm2e}

\newcounter{def_counter}
\newcounter{pro_counter}
\newtheorem{definition}[def_counter]{Definition}
\newtheorem{proposition}[pro_counter]{Proposition}

\makeatletter
\g@addto@macro\normalsize{%
\setlength\abovedisplayskip{5pt}
\setlength\belowdisplayskip{3pt}
\setlength\abovedisplayshortskip{5pt}
\setlength\belowdisplayshortskip{3pt}
}
\makeatother

\def\rb{\textcolor{black}}

\cvprfinalcopy % *** Uncomment this line for the final submission

\def\cvprPaperID{5642} % *** Enter the CVPR Paper ID here

\begin{document}

%=============================================
%=============================================
%=============================================

\title{Automatic Neural Network Compression by Sparsity-Quantization Joint Learning: A Constrained Optimization-based Approach}

\author[1]{Haichuan~Yang}
\author[1]{Shupeng~Gui}
\author[1]{Yuhao~Zhu}
\author[2]{Ji~Liu}

% \affil[]{\small Department of Computer Science and Department of Electrical and Computer Engineering, \\University of Rochester, \\Rochester, NY 14627, USA}

\affil[1]{Department of Computer Science, University of Rochester, Rochester, USA}
\affil[2]{AI Platform, Ytech Seattle AI Lab, FeDA Lab, Kwai Inc., Seattle, USA}
\maketitle

\begin{abstract}
Deep Neural Networks (DNNs) are applied in a wide range of usecases. There is an increased demand for deploying DNNs on devices that do not have abundant resources such as memory and computation units. Recently, network compression through a variety of techniques such as pruning and quantization have been proposed to reduce the resource requirement. A key parameter that all existing compression techniques are sensitive to is the compression ratio (e.g., pruning sparsity, quantization bitwidth) of each layer. Traditional solutions treat the compression ratios of each layer as hyper-parameters, and tune them using human heuristic. Recent researchers start using black-box hyper-parameter optimizations, but they will introduce new hyper-parameters and have efficiency issue. In this paper, we propose a framework to jointly prune and quantize the DNNs \textit{automatically} according to a target model size without using any hyper-parameters to manually set the compression ratio for each layer. In the experiments, we show that our framework can compress the weights data of ResNet-50 to be 836$\times$ smaller without accuracy loss on CIFAR-10, and compress AlexNet to be 205$\times$ smaller without accuracy loss on ImageNet classification.
\end{abstract}

\section{Introduction}

\label{sec:intro}
Nowadays, Deep Neural Networks (DNNs) are being applied everywhere around us. Besides running inference tasks on cloud servers, DNNs are also increasingly deployed in resource-constrained environments today, ranging from embedded systems in micro aerial vehicle and autonomous cars to mobile devices such as smartphones and Augmented Reality headsets. In these environments, DNNs often operate under a specific resource constraint such as the model size, execution latency, and energy consumption. Therefore, it is critical to compress DNNs to run inference under given resource constraints while maximizing the accuracy.

In the past few years, various techniques have been proposed to compress the DNN models. Pruning and quantization are two of which most widely used in practice. Pruning demands the weights tensor to be sparse, and quantization enforces each DNN weight has a low-bits representation. These methods will compress the DNN weights in each layer and result in a compressed DNN having lower resource consumption. It has been shown that by appropriately 
%setting the compression ratio (i.e., sparsity or quantization bitwidth) for each layer, 
setting the compression rate and performing fine-tuning,
the compression could bring negligible accuracy drop~\citep{han2015deep}.

\begin{table}[tb]
	\Huge
	\centering
	\caption{\small Comparison across different \emph{automated} model compression methods.}
	\renewcommand*{\arraystretch}{1.1}
	\renewcommand*{\tabcolsep}{10pt}
	\resizebox{\columnwidth}{!}
	{
		\begin{tabular}{c|c|c|c|c}
			\toprule[0.15em]
			Methods $\backslash$ Features                              & Pruning & Quantization & Automated & End-to-end\\
			\midrule[0.05em]
			%ECC~\citep{yang2018end} & \checkmark &  & \checkmark  \\
			%\citet{yang2018ecc} & \checkmark &  & \checkmark  \\
			AMC~\citep{he2018amc} & \cmark & &\cmark &  \\
			HAQ~\citep{wang2019haq} &  & \cmark &\cmark & \\
			CLIP-Q~\citep{tung2018clip}  &\cmark &\cmark & \cmark & \\
			\hline
			\textbf{Ours} & \cmark & \cmark &\cmark &\cmark \\
			\bottomrule[0.15em]
		\end{tabular}
	}
	\label{tab:comparison}
\end{table}

%Recent researches~\citep{liu2018rethinking, frankle2018lottery} find that the performance of the compressed DNN is largely determined by the pruned architecture, which in turn is determined by the layer-wise sparsity such as ~\citep{he2018amc, yang2018netadapt}.

\begin{figure*}[t]
    \centering
    \includegraphics[width=0.8\textwidth]{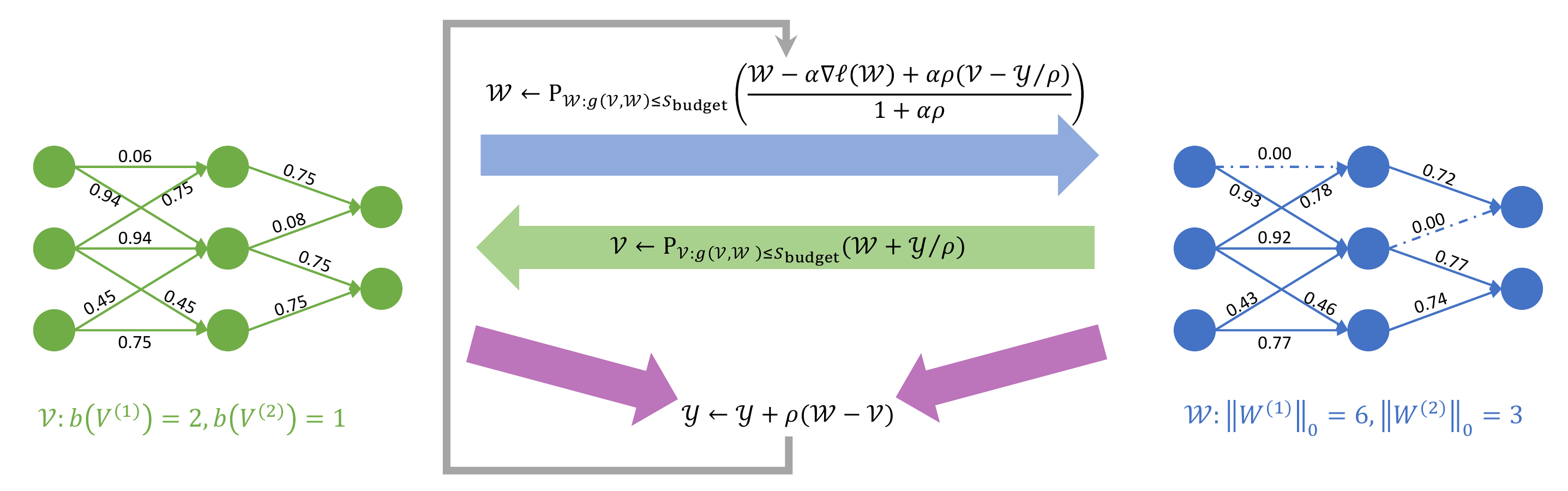}
    \caption{Illustration of the proposed DNN compression framework. DNN weight $\cW$ is sparse and $\cV$ is quantized. $\cV$ is a ``soft duplicate'' of $\cW$ and they are converged to be equal.}
    \label{fig:view}
    	\vspace{-8pt}
\end{figure*}

Recent research works~\citep{yang2018netadapt, he2018amc, wang2019haq, liu2019metapruning} \rb{found} that given the resource constraint, the accuracy of compressed DNNs can be further improved by tuning the compression ratio (i.e., sparsity or quantization bitwidth) for each layer.
A fundamental question is: \textit{how to find the optimal compression ratio, e.g., sparsity and/or bitwidth, for each layer in a way that meets a given resource budget}. Traditional DNN compression methods~\citep{han2015deep, ye2018unified, he2019filter} set the compression ratio of each layer based on human heuristics. Since the compression ratios can be seen as hyper-parameters, the idea in recent research of using black-box optimization for hyper-parameter search can be directly adopted~\citep{tung2018clip}. He~et~al.~\citet{he2018amc} \rb{applied} reinforcement learning (RL) in DNN pruning by formulating the pruning ratio as a continuous action and the accuracy as the reward. Wang~et~al.~\citet{wang2019haq} \rb{applied} the similar formulation but \rb{used} it for searching the quantization bitwidth of each layer. CLIP-Q~\citep{tung2018clip} \rb{proposed} a compression method which \rb{required} the sparsity and quantization bitwidth to be set as hyper-parameters, and they \rb{used} Bayesian optimization libraries to search them. Evolutionary search (ES) \rb{was} also being used in this scenario, for example,
%Guo~et~al.~\citet{guo2019single} propose an ES-based network architecture search (NAS) method and use it for searching compression ratios.
Liu~et~al.~\citet{liu2019metapruning} \rb{used} meta-learning and ES to find the pruning ratios of channel pruning. The basic idea of these methods \rb{was} formulating the compression ratio search as a black-box optimization problem, but it \rb{introduced} new hyper-parameters in the RL or ES algorithm. However, tuning black-box optimization algorithms could be very tricky~\citep{islam2017reproducibility} and usually inefficient~\citep{rlblogpost}. Moreover, it introduces new hyper-parameters. For example, the RL algorithm DDPG~\citep{lillicrap2015continuous} \rb{had} dozens of hyper-parameters including batch size, actor / critic network architecture, actor /critic optimizer and learning rate, reward scale, discounting factor, reply buffer size, target network updating factor, exploration noise variance, and so on. Therefore, it is highly desirable to have an automated approach avoiding as much as possible the human heuristics.

% Meanwhile, to maximize the compression performance pruning and quantization could be used together~\citep{han2015deep}. Under this circumstance, a compression algorithm must tune both sparsity and quantization bitwidth for each layer. These two parameters will influence each other. For example, if layer $i$ has higher bitwidth than another layer $j$, then pruning layer $i$ (i.e., reducing the number of nonzero elements) will contribute more to model compression than pruning layer $j$. So jointly pruning and quantization increases the difficulty of manually choosing the compression ratios or hyper-parameter tuning.

Meanwhile, to maximize the compression performance, pruning and quantization could be applied simultaneously~\citep{han2015deep}. Thus, the layer-wise sparsity and quantization bitwidth will affect each other under this circumstance. For example, if layer $i$ has larger bitwidth than layer $j$, then pruning layer $i$ will contribute more than pruning layer $j$. Joint pruning and quantization increase the difficulty of manually choosing the compression ratios or hyper-parameter tuning.

%Thus, most existing methods focus only on one compression technique, and there is no good automated solution that simultaneously applies both techniques for compression, to our best knowledge.

In this paper, we present an end-to-end framework for automatic DNN compression. Our method can jointly prune and quantize the DNN model, and simultaneously learn the compression ratios and the compressed model weights. Instead of treating the compression ratios as hyper-parameters and using the black-box optimization, our method is based on a constrained optimization where an overall model size is set as the constraint to restrict the structure of the compressed model weights. Table~\ref{tab:comparison} shows a comparison of our method with recently proposed automated model compression works.

The main contributions of this paper are summarized as follows:
\begin{itemize}[itemsep=0pt,topsep=0pt,leftmargin=*]
    %\item We propose an end-to-end framework to automatically compress DNNs without manually setting the compression ratio for each layer. A compression budget could be set by users to prune and quantize a compressed DNN simultaneously where the DNN can have different sparsity and bitwidth for each layer.
    \item We propose an end-to-end framework to automatically compress DNNs without manually setting the compression ratio for each layer. It allows the user to set a budget and simultaneously utilizes pruning and quantization.
    % and directly learns the compressed DNN weights which can have different sparsity and bitwidth for each layer.
    \item We mathematically formulate the automated compression problem to a constrained optimization problem. The problem has a ``sparse + quantized'' constraint and it is further decoupled so that we can solve it using the Alternating Direction Method of Multipliers (ADMM)~\citep{boyd2011distributed}.
    %\item A mathematical formulation of the compression problem is proposed which contains a ``sparse + quantized'' constraint. And we can solve this problem with the help of Alternating Direction Method of Multipliers (ADMM)~\citep{boyd2011distributed}.
    \item The main challenge in using ADMM for the automated compression problem is solving the projection operators for pruning and quantization. We introduce the algorithms for getting the projection of the sparse constraint and quantization constraint. In the experiment, we validate our automated compression framework to show its superiority over the handcrafted and black-box hyper-parameter search methods.
\end{itemize}

\section{Related Work}
\label{sec:related}

%In this section, we review the related works form several categories.

\subsection{Model Compression Techniques}
Due to the enormous impactions of mobile computing, more and more complicated DNN models are required to fit into those low-power consumption devices for real application. To solve the computation consumption issue onto the mobile systems, pruning and quantization are proposed as two practical approaches nowadays. 

\paragraph{Pruning}
Pruning refers to decrease the amount of non-zero parameters in DNN models.
Han~et~al.~\citet{han2015learning} proposed a simple approach by zeroing out the weights whose magnitudes are smaller than a threshold. By performing fine-tuning after removing the smaller weights, the accuracy drop is usually negligible even with a considerable compression ratio~\citep{han2015deep}.
%Usually, there are plenty of non-significant (close to zero) parameters in layers of DNN which have very small impact to the performance of the model. Hence, pruning sparsifies those weights by replacing them with zeros~\citep{han2015learning}.
%Various types of regularizations are applied to decrease the size of the models in training process such as \citep{liu2015sparse, zhou2016less}.
Besides using weights pruning for model compression, channel (filter / neuron) pruning~\citep{li2016pruning, zhou2016less, molchanov2016pruning, he2017channel, luo2017thinet, zhuang2018discrimination, liu2017learning, ye2018rethinking} \rb{was} proposed to remove the entire filter of the CNN weights, thus also \rb{achieved} inference acceleration.
%can be treated as a special case of pruning with a structured pattern in weights of models. 
Wen~et~al.~\citet{wen2016learning} introduced more sparsity structures into CNN pruning, such as shape-wise and depth-wise sparsity.
%\citet{he2018progressive} proposed a progressive soft filter pruning method (PSFP) to prune the filters of convolutional layers progressively, which relieved the impact of pre-trained sparse models. 
%\citet{molchanov2016pruning} discussed a pruning criterion with respect to the approximation of change on the cost function caused by the pruned parameters in DNN models for transfer learning. ThiNet~\citep{luo2017thinet} claimed that the filter pruning should rely on the statistics information of the next layer, not the current one and it introduced a formal optimization problem to prune the filters.
%\citet{hu2016network} proposed a network trimming method which prunes the neurons with mostly zero output no matter what inputs the network received. They measured the neuron importance with average percentage of zero outputs.
%\citet{he2017channel} presented a pruning method which consists of two steps: 1) a LASSO regression based channel selection and 2) least square reconstruction.
%\citet{anwar2016compact} figured out a strategy that combined the feature map and kernel level pruning at the same time. Feature map pruning helps to reduce the width of DNN and the kernel pruning further reduces the model size. 
%\citet{ye2018rethinking} executed channel pruning with a two-step strategy. 

\paragraph{Quantization}
Besides decreasing the number of parameters with pruning, quantization is considered as another direction to compress DNNs.
To relieve the cost of memory storage or computation, quantization focuses on converting the floating-point number elements to low-bits representations.
For example, \rb{one} can quantize all the parameters' precision from 32 bits to 8 bits or lower~\citep{han2015deep} to down-scale the model size. Extremely, the model weights can be binary~\citep{courbariaux2015binaryconnect, rastegari2016xnor, courbariaux2016binarized, hubara2017quantized}, or ternary~\citep{li2016ternary, zhu2016trained}.
The quantization interval can be either uniform~\citep{jacob2018quantization} or nonuniform~\citep{han2015deep, miyashita2016convolutional, tang2017train, zhang2018lq}. Typically, nonuniform quantization can achieve higher compression rate, while uniform quantization can provide acceleration. The quantization bitwidth could be further reduced by Hoffman coding~\citep{han2015deep, choi2016towards}.
%Additionally, \citet{wang2018two} proposed a two-step quantization method TSQ which further improved the quantization performance with fixed quantization bits, binary or ternary. 
%\citet{park2017weighted} quantized the network with arbitrary bits for both weights and activations and obtained high performance with low design effort. 
%\citet{xu2018deep} utilized different quantization methods (SLQ and MLQ) for various quantization bits.
%\citet{polino2018model} leveraged distillation with a teacher-student network training strategy and quantized the models with locating the quantization points through stochastic gradient descent.
Besides the scalar quantization, vector quantization was also applied in DNN model compression~\citep{gong2014compressing, wu2018deep}.

There are some methods \rb{performing} training together with pruning and quantization, including Ye~et~al.~\citet{ye2018unified} and CLIP-Q~\citep{tung2018clip}. These methods \rb{relied} on setting hyper-parameters to compress the layers with desired compression ratios, though the black-box hyper-parameter optimization method can be used~\citep{tung2018clip}.
Recently, ADMM was used to formulate and solve model compression problems~\citep{leng2018extremely, zhang2018systematic, ye2019progressive, ren2019admm, gui2019model}.
%However, our method has a different focus from these prior methods.
However, These prior methods require the per-layer sparsity / bitwidth to be manually set. The main contribution of this paper is presenting an end-to-end framework to automatically prune and quantize DNNs without manually setting the compression ratio for each layer.
%These methods~\citep{leng2018extremely, zhang2018systematic, ye2019progressive, ren2019admm} require the per-layer pruning ratio / quantization bitwidth to be manually set as hyper-parameters.

\subsection{Automated Model Compression}
Prior efforts on setting for the compression ratio of each layer mostly \rb{used} either rule-based approaches~\citep{han2015deep, howard2017mobilenets, ye2018unified, he2019filter} or black-box hyper-parameter search. Rule-based approaches \rb{relied} on heuristics, and thus \rb{were} not optimal and unscalable as network architectures becoming more complex. Search-based approaches \rb{treated} this problem as hyper-parameter search to eliminate the need for human labor. For pruning, NetAdapt~\citep{yang2018netadapt} \rb{applied} a greedy search strategy to find the sparsity ratio of each layer by gradually decreasing the resource budget and performing fine-tuning and evaluation iteratively. In each iteration, NetAdapt \rb{tried} to reduce the number of nonzero channels of each layer, and \rb{picked} the layer which results in smallest accuracy drop. Recent search-based approaches also \rb{employed} reinforcement learning (RL), which \rb{used} the accuracy and resource consumption to define the reward and guide the search to find pruning ratio~\citep{he2018amc} and quantization bitwidth~\citep{yazdanbakhsh2018releq, wang2019haq}. Guo~et~al.~\citet{guo2019single} \rb{used} evolutionary search (ES) for network architecture search (NAS) and \rb{showed} that it could be used for searching compression ratios. Liu~et~al.~\citet{liu2019metapruning} \rb{used} a hyper-network in the ES algorithm to find the layer-wise sparsity for channel pruning. Instead of regarding the layer-wise sparsity as hyper-parameters, recently proposed energy-constrained compression methods~\citep{yang2018ecc, yang2018end} \rb{used} optimization-based approaches to prune the DNNs under a given energy budget. Besides the above, there are some methods on searching efficient neural architectures~\citep{cai2018proxylessnas, tan2019mnasnet}, while our work mainly concentrates on compressing a given architecture.

\section{End-to-end Automated DNN Compression}
\label{sec:annc}

In this section, we firstly introduce a general formulation of DNN compression, which is constrained by the total size of the compressed DNN weights.
%Then we propose our DNN compression framework with joint pruning and quantization.
Secondly, we reformulate the original constraint to decouple the pruning and quantization and show the algorithm outline which uses ADMM to solve the constrained optimization. Lastly, as the proposed algorithm requires two crucial projection operators, we show that they can be formed as special integer linear programming (ILP) problems and introduce efficient algorithms to solve them.

\subsection{Problem Formulation}
\label{sec:formulation}
Let $\cW := \{W^{(i)}\}_{i=1}^L$ be the set of weight tensors of a DNN which has $L$ layers. To learn a compressed DNN having a target size of $S_{\text{budget}}$, we have the constrained problem
\begin{equation}
\min_{\cW} \ \ell(\cW), \quad \text{s.t. } \sum_{i=1}^L b(W^{(i)}) \|W^{(i)}\|_0 \leq S_{\text{budget}}, \label{eq:obj1}
\end{equation}
where $b(W)$ is the minimum bitwidth to encode all the nonzero elements of tensor $W$, 
%which means the number of unique nonzero values in $W$ is at most $2^{b(W)}$,
i.e., $b(W)=\lceil \log_2|\{\text{unique nonzero elements of $W$}\}| \rceil$. $L_0$-norm $\|W\|_0$ is the number of nonzero elements of $W$. The loss function $\ell$ is task-driven, for example, using the cross entropy loss as $\ell$ for classification, or mean squared error for regression.

Problem~\eqref{eq:obj1} is a general form of DNN compression. When assuming the bitwidth is fixed and same for all the layers, problem~\eqref{eq:obj1} reduces to the case of weights pruning~\citep{han2015learning}. When assuming the weight tensors are always dense, it is reduced to mixed-bitwidth quantization~\citep{wang2019haq}.

Compared with the ordinary training of deep learning, the compressed DNN learning problem~\eqref{eq:obj1} introduces a constraint, i.e. $\sum_{i=1}^L b(W^{(i)}) \|W^{(i)}\|_0 \leq S_{\text{budget}}$. It is defined by two non-differentiable functions $b(\cdot)$ and $\|\cdot\|_0$, which obstruct solving it via normal training algorithm. Although there is a projection-based algorithm which can handle the $L_0$-norm constraint, it can not be applied to our case because our constraint sums the products of $\|\cdot\|_0$ and $b(\cdot)$, which is more complicated.

\subsection{Constraint Decoupling via Alternating Direction Method of Multipliers}
We deal with the constraint in~\eqref{eq:obj1} by decoupling its $L_0$-norm and bitwidth parts. Specifically, we reformulate the problem~\eqref{eq:obj1} to an equivalent form
\begin{equation}
\min_{\cW, \cV} \ell(\cW),\quad \text{s.t. } \cV = \cW,\ g(\cV, \cW) \leq S_{\text{budget}}.\label{eq:obj2}
\end{equation}
Where $\cV:=\{V^{(i)}\}_{i=1}^L$ is a duplicate of the DNN weights $\cW$, and $g(\cV, \cW):=\sum_{i=1}^L b(V^{(i)}) \|W^{(i)}\|_0$.

In this paper, we apply the idea from ADMM to solve the above problem. We introduce the dual variable $\cY:=\{Y^{(i)}\}_{i=1}^L$ and absorb the equality constraint into the augmented Lagrangian $\cL_{\rho}(\cW, \cV,\cY):=\ell(\cW) + \langle \cY ,\cW - \cV \rangle + (\rho/2)\|\cW - \cV \|^2$, i.e.,
\begin{subequations}
\begin{align}
\min_{\cW, \cV} \max_{\cY} \ &\ell(\cW) + \langle \cY ,\cW - \cV \rangle + \frac{\rho}{2}\|\cW - \cV \|^2, \\
%\min_{\cW, \cV} \max_{\cY} &\ \cL_{\rho}(\cW, \cV,\cY) := \ell(\cW) + \sum_{i=1}^L \langle Y^{(i)},W^{(i)} - V^{(i)}\rangle + (\rho/2)\sum_{i=1}^{L}\|W^{(i)} - V^{(i)}\|^2, \\
\text{s.t. } & g(\cV, \cW) \leq S_{\text{budget}},
\end{align}
\label{eq:obj3}
\end{subequations}
where $\rho>0$ is a hyper-parameter.
Based on ADMM, we can solve this problem by updating $\cW, \cV$ and $\cY$ iteratively. In each iteration $t$, we have three steps corresponding to the variable $\cW, \cV$ and $\cY$ respectively.

\paragraph{Fix $\cV, \cY$, update $\cW$.} In this step, we treat $\cV, \cY$ as constants and update $\cW$ to minimize $\cL_{\rho}$, i.e., $\cW^{t+1} =$
\begin{align}
&\argmin_{\cW: g(\cV^t, \cW) \leq S_{\text{budget}}} \ell(\cW) + \langle \cY^t, \cW -\cV^t \rangle + \frac{\rho}{2}\|\cW - \cV^t \|^2 \notag\\
& = \argmin_{\cW: g(\cV^t, \cW) \leq S_{\text{budget}}} \ell(\cW) + \frac{\rho}{2} \| \cW - \cV^t + {1\over \rho}\cY^t \|^2 \label{eq:wobj1}.
 \end{align}
Because of the complexity of the DNN model and the large amount of the training data, $\ell(\cdot)$ is usually complex and the gradient based algorithms are often used to iteratively solve it. To support the gradient-based updating, we apply a proximal gradient method. Specifically, the loss function $\ell(\cW)$ is substituted with its first-order expansion, i.e., the problem~\eqref{eq:wobj1} becomes
%Here we apply a proximal method to simplify the objective~\eqref{eq:wobj1}. Firstly, use a quadratic proxy to approximate $\ell(\cW)$, the problem~\eqref{eq:wobj1} becomes
{\small
\begin{align}
\argmin_{\cW: g(\cV^t, \cW) \leq S_{\text{budget}}} &\ell(\cW^t) + \langle \nabla \ell(\cW^t), \cW - \cW^t \rangle \notag\\
&+ {1\over 2\alpha}\| \cW - \cW^t \|^2
+ {\rho\over 2}\| \cW -\cV^t + {1\over \rho}\cY^t \|^2 \notag\\
%=& \argmin_{\cW} \| \cW - (\cW^t - \alpha\nabla \ell(\cW^t)) \|^2 + \alpha \rho \| \cW - ({\cV^t} - {1\over \rho}{\cY^t}) \|^2 \notag\\
= \argmin_{\cW: g(\cV^t, \cW) \leq S_{\text{budget}}}& \left\| \cW - \bar{\cW} \right\|^2. \label{eq:wobj2}
%& \text{s.t. } \sum_{i=1}^L b({V^{t}}^{(i)}) \|W^{(i)}\|_0 \leq S_{\text{budget}}.
\end{align}}
Where $\bar{\cW}:={1\over 1 + \alpha \rho} (\cW^t - \alpha\nabla \ell(\cW^t) + \alpha\rho ({\cV^t} - {1\over \rho}{\cY^t}))$, $\nabla \ell(\cW^t)$ is the (stochastic) gradient of $\ell$ at point $\cW^t$, $\alpha$ is the learning rate, and .
%Let $\Phi(\cV^t,S_{\text{budget}}):=\{\cW: \sum_{i=1}^L b({V^{t}}^{(i)}) \|W^{(i)}\|_0 \leq S_{\text{budget}}\}$ denote the constraint set, 
Problem~\eqref{eq:wobj2} is the projection of $(\cW^t - \alpha\nabla \ell(\cW^t) + \alpha\rho ({\cV^t} - {1\over \rho}{\cY^t}))/(1 + \alpha \rho)$ onto the set $\{\cW:g(\cV^t, \cW) \leq S_{\text{budget}}\}$. We call it the compression projection with fixed bitwidth, and show how to solve it in Section~\ref{sec:proj1}.

\paragraph{Fix $\cW, \cY$, update $\cV$.} Here we use the updated $\cW^{t+1}$ and minimize $\cL_\rho$ in terms of $\cV$.
\begin{equation}
\cV^{t+1} = \argmin_{\cV: g(\cV, \cW^{t+1}) \leq S_{\text{budget}}} \| {\cW^{t+1}} - {\cV} + {1\over \rho}{\cY^t} \|^2.
\label{eq:vobj}
\end{equation}
Since $\cW^{t+1}$ and $\cY^t$ are fixed in this step, they can be seen as constants here.
%Let $\Psi(\cW^{t+1},S_{\text{budget}}):=\{\cV : \sum_{i=1}^L b({V}^{(i)}) \|{W^{t+1}}^{(i)}\|_0 \leq S_{\text{budget}}\}$ be the constraint set, 
Problem~\eqref{eq:vobj} is the projection of $\cW^{t+1} + {1\over \rho}{\cY^t}$ onto $\{\cV:g(\cV, \cW^{t+1}) \leq S_{\text{budget}}\}$. We call this projection the compression projection with fixed sparsity and leave the detail of solving it in Section~\ref{sec:proj2}.

\paragraph{Fix $\cW, \cV$, update $\cY$.} To update the dual variable $\cY$, we perform a gradient ascent step with learning rate as $\rho$:
\begin{equation}
\cY^{t+1} = \cY^{t} + \rho (\cW^{t+1} - \cV^{t+1}).
\end{equation}

The above updating rules follow the standard ADMM.
%Although ADMM relies on several assumptions (e.g., the objective function should be convex), 
Recent theoretical analysis shows the convergence of ADMM also holds on non-convex problems~\citep{wang2019global}.
%\todo{ADMM non-convex theory citation\cite{wang2019global}} we apply it in non-convex loss function and non-differentiable constraint functions to solve the minimax problem~\eqref{eq:obj3}.
In Section~\ref{sec:exp}, we demonstrate these updating rules work well in our problem.

\subsection{Compression Projection with Fixed Bitwidth}
\label{sec:proj1}
Problem~\eqref{eq:wobj2} can be seen as a weighted $L_0$-norm projection $\proj_{\cW: g(\cV^t, \cW) \leq S_{\text{budget}}}(\bar{\cW})$ with $\bar{\cW}=(\cW^t - \alpha\nabla \ell(\cW^t) + \alpha\rho ({\cV^t} - {1\over \rho}{\cY^t}))/(1 + \alpha \rho)$:
\begin{align}
\proj_{\cW : g(\cV^t, \cW) \leq S_{\text{budget}}}(\bar{\cW}) := &\argmin_{\cW} \|\cW - \bar{\cW}\|^2, \label{eq:proj1}\\
 \text{s.t. } &\sum_{i=1}^L b({V^{t}}^{(i)}) \|W^{(i)}\|_0 \leq S_{\text{budget}} \notag. 
\end{align}

We will show that this is actually a 0-1 Knapsack problem~\citep{yang2018energy}.
\begin{proposition} \label{prop:reformulation}
The projection problem in \eqref{eq:proj1} is equivalent to the following 0-1 Knapsack problem:
\begin{equation}
\label{eq:knapsack}
\max_{\cX~\text{\rm is binary}} \langle \bar{\cW}^2, \cX \rangle, \quad \text{\rm s.t. } \langle \cA, \cX \rangle \leq S_{\text{budget}}, %\notag
\end{equation}
where $\cA$ and $\cX$ are of the same shape as $\bar{\cW}$, and the elements of $A^{(i)}$ is defined as $A^{(i)}_j = b({V^{t}}^{(i)}),\ \forall j$. $\bar{\cW}^2$ takes element-wise square of $\bar{\cW}$. The optimal solution of~\eqref{eq:proj1} is $\proj_{\cW : g(\cV^t, \cW) \leq S_{\text{budget}}}(\bar{\cW})=\cX^* \odot \bar{\cW}$, where $\cX^*$ is the optimal solution to the knapsack problem~\eqref{eq:knapsack} and $\odot$ is the element-wise multiplication.
\end{proposition}

In this 0-1 Knapsack problem, $\bar{\cW}^2$ is called the ``profit'', and $\cA$ is the ``weight''. The 0-1 Knapsack is basically selecting a subset of items (corresponding to the DNN weights in our case) to maximize the sum of the profit and the total weight does not exceed the budget $S_{\text{budget}}$. The 0-1 Knapsack problem is NP hard, while there exists an efficient greedy algorithm~\citep{hans2004knapsack} which works well in practice. The idea is based on the profit to weight ratio $({\bar{W}}^{(i)}_j)^2 / A_j^{(i)}$. We sort all items based on this ratio and iteratively select the largest ones until the constraint boundary is reached. The theoretical complexity of this algorithm is $O(n\log(n))$, where $n$ is the number of total items. Because the sorting and cumulative sum operations are supported on GPU, we can efficiently implement this algorithm on GPU and use it in our DNN compression framework.

\subsection{Compression Projection with Fixed Sparsity}
\label{sec:proj2}
The solution of problem~\eqref{eq:vobj} is the projection $\proj_{\cV: g(\cV, \cW^{t+1}) \leq S_{\text{budget}}}(\cW^{t+1} + {1\over \rho}{\cY^t})$, where the projection operator $\proj_{\cV: g(\cV, \cW^{t+1}) \leq S_{\text{budget}}}(\cdot)$ is defined as
\begin{align}
\proj_{\cV: g(\cV, \cW^{t+1}) \leq S_{\text{budget}}}(\bar{\cV}) := &\argmin_{\cV} \| {\cV} - \bar{\cV}\|^2, \label{eq:proj2}\\
\textrm{s.t. } &\sum_{i=1}^L b({V}^{(i)}) \|{W^{t+1}}^{(i)}\|_0 \leq S_{\text{budget}}.\notag
\end{align}

The above problem can be also reformulate as an integer linear programming. In the following, we will introduce a special variant of Knapsack problem called Multiple-Choice Knapsack Problem (MCKP)~\citep{hans2004knapsack}  and show that the problem~\eqref{eq:proj2} can be written as an MCKP.

\begin{definition}
\label{def:mckp}
Multiple-Choice Knapsack Problem (MCKP)~\citep{hans2004knapsack}. Consider there are $L$ mutually disjoint groups $G_1,...,G_L$ which contain $n_1,...,n_L$ items respectively. The $j$-th item from the $i$-th group has a ``profit'' $\rho_{ij}$, and ``weight'' $\omega_{ij}$, $\forall i=1,...,L, j\in 1,...,n_i$. MCKP formulates how to select exactly one item from each group to maximize the sum of profits and keep the sum of weights under a given budge $\beta$, i.e.,
\begin{subequations}
\begin{align*}
&\max_{\x~\text{\rm is binary}} \sum_{i=1}^L \sum_{j=1}^{n_i} \rho_{ij}\x_{ij},\\
%\text{\rm s.t. } & \x_{ij} \in \{0, 1\}, \forall i=1,...,L, j=1,...,n_i; \\
& \textrm{s.t. }  \sum_{j=1}^{n_i} \x_{ij} = 1, \forall i=1,...,L; \ \sum_{i=1}^L \sum_{j=1}^{n_i} \omega_{ij}\x_{ij} \leq \beta.
\end{align*}
\end{subequations}

\end{definition}

Define $\cB$ as the set of bitwidth candidates. In this paper, we use $\cB=\{1,2,3,...,8\}$. Let $\cE_j(\bar{V})$ be the error to quantize $\bar{V}$ with bitwidth $j$, i.e., $\cE_j(\bar{V}) = \min_{V: b(V)=j} \|V -\bar{V} \|^2$, which can be solved by k-means algorithm for nonuniform quantization~\citep{han2015deep}. Now we are ready to reformulate the problem~\eqref{eq:proj2} as an MCKP.
\begin{proposition}
\label{prop:mckp}
The compression projection problem~\eqref{eq:proj2} can be reformulated to an instance of MCKP in Definition~\ref{def:mckp}. Specifically, each group $G_i$ is defined by each layer and has size $n_i = |\cB|$. Each choice of the quantization bitwidth is regraded as an MCKP item. The profit $\rho_{ij}$ is $-\cE_j(\bar{V}^{(i)})$, the weight $\omega_{ij}$ is $j\|{W^{t+1}}^{(i)}\|_0$, the Knapsack budget $\beta$ is $S_{\text{budget}}$, and $\x_{ij}$ indicates selecting which bitwidth.
\end{proposition}

The MCKP is also NP-hard. However, if we relax the binary constraints $\x_{ij} \in \{0,1\}$ to $\x_{ij} \in [0,1]$, it is reduced to a Linear Programming and can be solved efficiently. \citet{Zemel} transforms the linear relaxation of MCKP to the fractional knapsack problem and use a greedy algorithm to solve it. Based on this idea, we can get a feasible MCKP solution by the following steps: %in~\citet[Fig. 11.2]{hans2004knapsack}:
\begin{enumerate}[itemsep=0pt,topsep=0pt,leftmargin=*]
\item For each group, sort the items based on their weights in ascending order, i.e., $\omega_{ij'} \geq \omega_{ij}$ if $j' \geq j$. According to ~\citet[Proposition 11.2.2]{hans2004knapsack}, the profits of the sorted items are nondecreasing, i.e., $\rho_{ij'} \geq \rho_{ij}$ if $\omega_{ij'} \geq \omega_{ij}$. The incremental profit density $(\rho_{ij} - \rho_{i,j-1}) / (\omega_{ij} - \omega_{i,j-1})$ has descending order, i.e., $(\rho_{ij'} - \rho_{i,j'-1})/(\omega_{ij'} - \omega_{i,j'-1}) \leq (\rho_{ij} - \rho_{i,j-1}) / (\omega_{ij} - \omega_{i,j-1})$ if $\omega_{ij'} \geq \omega_{ij}$.
\item Select the first item (having the smallest weight) of each group. It should be noted that the budget must be large enough to contain these items, otherwise there is no feasible solution under the constraints.
\item For other items, select the one with the largest incremental profit density. When selecting the $j$-th item of the $i$-th group, discard the $(j-1)$-th item. Repeat the same procedure for the 2nd, 3rd, ... largest ones, until the total weight of selected items exceeds the budget.
\end{enumerate}

The above algorithm can find a feasible MCKP solution, i.e., selecting one item from each group and guarantee their total weight is under the given budget $\beta$. Its time complexity is $O(L|\cB|\log(L|\cB|))$. In practice, $L$ and $|\cB|$ are much smaller than the number of DNN weights, so the time complexity of this algorithm is negligible.
%Although this greedy solution is not always global optimal, it
The greedy solution has some nice properties and could be global optimal in some cases~\citep[Corollary 11.2.3]{hans2004knapsack}. By using the above algorithm to solve our compression projection problem~\eqref{eq:proj2}, we can get the projection result of $\proj_{\cV: g(\cV, \cW^{t+1}) \leq S_{\text{budget}}}(\cdot)$, which essentially allocates the bitwidth across different layers.

We summarize the training procedure of our method in Algorithm~\ref{alg:main}. We use $\tau$ to denote the number of total SGD iterations of our algorithm. For large scale datasets, the number of SGD iterations could be very large. So we do not make the projections and dual update every time after we perform the proximal SGD on $\cW$, but use a hyper-parameter $\tau'$ to control the frequency of dual updates. $\tau$ should be divisible by $\tau'$. In our experiments, $\tau'$ is set to be the iteration number of one epoch, since we do not observe any improvement by using smaller $\tau'$.
%Ideally, we should get $\cW$ converged to $\cV$ in the end, but t is hard to get $\cW$ exactly equals to $\cV$ in practice. So we perform a quantization to $\cW$ to guarantee it satisfies the model size constraint.

\begin{algorithm2e}[htbp]
 \SetAlgoLined
% \KwIn{Original DNN parameterized $\cW$, compression budget $S_{\text{budget}}$, primal learning rate $\alpha$, dual learning rate $\rho$, iteration number $\tau$, and the dual updating interval $\tau'$.}
  \KwIn{Original DNN parameterized $\cW$, compression budget $S_{\text{budget}}$.}
 \KwResult{The compressed DNN weights $\cW^*$.}
 Initialize $\cW$ with pretrained dense model, initialize $\cV$ by uniformly quantizing $\cW$, and initialize $\cY=\0$;\\
 $\cW \leftarrow \proj_{\cW: g(\cV, \cW) \leq S_{\text{budget}})}(\cW)$;\\
 $\cV \leftarrow \proj_{\cV: g(\cV, \cW) \leq S_{\text{budget}}}(\cW + {1\over \rho}{\cY})$;\\
$\cY \leftarrow \cY + \rho (\cW - \cV)$;\\
 \For{$t \leftarrow 1$ \KwTo $\tau$}{
	Compute stochastic gradient $\nabla \ell(\cW)$;
	$\cW \leftarrow (\cW - \alpha\nabla \ell(\cW) + \alpha\rho ({\cV} - {1\over \rho}{\cY}))/(1 + \alpha \rho)$;\\
 	\If{$t \pmod {\tau' }= 0$}{
		$\cW \leftarrow \proj_{\cW: g(\cV, \cW) \leq S_{\text{budget}})}(\cW)$;\\
		$\cV \leftarrow \proj_{\cV: g(\cV, \cW) \leq S_{\text{budget}}}(\cW + {1\over \rho}{\cY})$;\\
		$\cY \leftarrow \cY + \rho (\cW - \cV)$;
		}
 }
 
% \For{$i \leftarrow 1$ \KwTo $L$}{
% \tcc{In case $\cW$ has not converged to $\cV$ yet.}
%Quantize $W^{(i)}$ with the bitwidth $b(V^{(i)})$;
%}
$\cW^* = \cW$.
\caption{Automatic DNN Compression.}
\label{alg:main}
\end{algorithm2e}

\section{Experiments}
\label{sec:exp}
In this section, we will evaluate our automated compression framework. We start with introducing the experiment setup such as evaluation and implementation details, then we show the compression results of our framework and compare it with state-of-the-art methods.

\subsection{Experiment Setup}
\paragraph{Datasets} We evaluate our method on three datasets which are most commonly used in DNN compression: MNIST~\citep{lecun1998gradient}, CIFAR-10~\citep{krizhevsky2009learning}, and ImageNet~\citep{deng2009imagenet}. We use the standard training / testing data split and data preprocessing on all the three datasets. For ImageNet, we evaluate on the image classification task (1000 classes).

\paragraph{DNN models} We evaluate on a wide range of DNN models, which are also used in current state-of-the-art compression methods. On MNIST, we use the LeNet-5 as in~\citep{han2015deep}.
%\footnote{\url{https://github.com/BVLC/caffe/tree/master/examples/mnist}}.
It has two convolution layers followed by two fully connected layers. For CIFAR-10, we evaluate on ResNet-20 and ResNet-50~\citep{he2016deep} which have 20 and 50 layers respectively. For ImageNet, we use the AlexNet~\citep{krizhevsky2012imagenet} and the well-known compact model MobileNet~\citep{howard2017mobilenets}. In addition, we also investigate the compression performance of our method on the most recently proposed compact architectures MnasNet~\citep{tan2019mnasnet} and ProxylessNAS-mobile~\citep{cai2018proxylessnas}, which are searched by NAS algorithms.

\paragraph{Baselines and metric} We compare our method with current state-of-the-art model compression methods related to ours. These methods include
Recently proposed automated pruning methods AMC~\citep{he2018amc} and Constraint-Aware Compression~\citep{chen2018constraint};
Recently proposed automated quantization methods ReLeQ~\citep{yazdanbakhsh2018releq} and HAQ~\citep{wang2019haq};
Methods which adopt both pruning and quantization: Deep Compression~\citep{han2015deep}, Bayesian Compression~\citep{louizos2017bayesian}, Ye~et~al.~\citep{ye2018unified}, and CLIP-Q~\citep{tung2018clip}.

Please refer to Tables~\ref{tab:imagenet}, \ref{tab:mnist}, and \ref{tab:cifar} for more detailed features of these methods.
%automated pruning method AMC~\citep{he2018amc} and Constraint-Aware Compression~\citep{chen2018constraint}, automated quantization methods ReLeQ~\citep{yazdanbakhsh2018releq} and HAQ~\citep{wang2019haq}, and jointly pruning and quantization methods~\citep{han2015deep, louizos2017bayesian,tung2018clip,ye2018unified}.
Although there are some overhead of the sparse index, we use the size of the compressed weights data to compute the compression rate since different indexing techniques may introduce unfairness in the comparison.
%because it directly corresponds to $S_{\text{budget}}$. In addition, different indexing techniques may introduce unfairness in the comparison.

\paragraph{Implementation details} We set the batch size as 256 for AlexNet and LeNet-5, and use 128 batch size on ResNets and MobileNet. We use the momentum SGD to optimize $\ell(\cW)$. We use initial learning rate $\alpha$ is set to $0.01$ on AlexNet and MobileNet, and $0.1$ on LeNet-5 and ResNets. We use the cosine annealing strategy~\citep{loshchilov2016sgdr} to decay the learning rate. We set the hyper-parameter $\rho=0.05$ for all the experiments. To make a more clear comparison, the compression budget $S_{\text{budget}}$ is set to be close to or smaller than the compared methods. Training is performed for 120 epochs on MNIST and CIFAR-10 and 90 epochs on ImageNet. Fine-tuning~\citep{han2015deep} is used on ImageNet for 60 epochs.
%fair comparison.
To guarantee the final $\cW^*$ satisfies the model size constraint, we directly perform a quantization to $\cW^*$ with the bitwidth of $\cV$.

%
%\subsection{DNN Compression Results}
%\paragraph{Convergence and Sensitivity of $\rho$} Todo...
%\begin{figure}[htbp]
%	\centering
%         \begin{subfigure}{.49\columnwidth}
%             \includegraphics[width=\linewidth]{training_curve.eps}
%                     \subcaption{\small Training loss.}
%         \end{subfigure}
%         \hfill
%          \begin{subfigure}{.49\columnwidth}
%             \includegraphics[width=\linewidth]{residual.eps}
%          	\subcaption{\small $\text{MSE}(\cW, \cV)$.}
%         \end{subfigure}
%         \caption{Training loss and MSE (between $\cW$ and $\cV$) with different values of $\rho$.}
%	\label{fig:convergence}
%\end{figure}
%

\begin{figure*}[htbp]
	\centering
	    \begin{subfigure}{0.63\columnwidth}
	        \includegraphics[width=\linewidth]{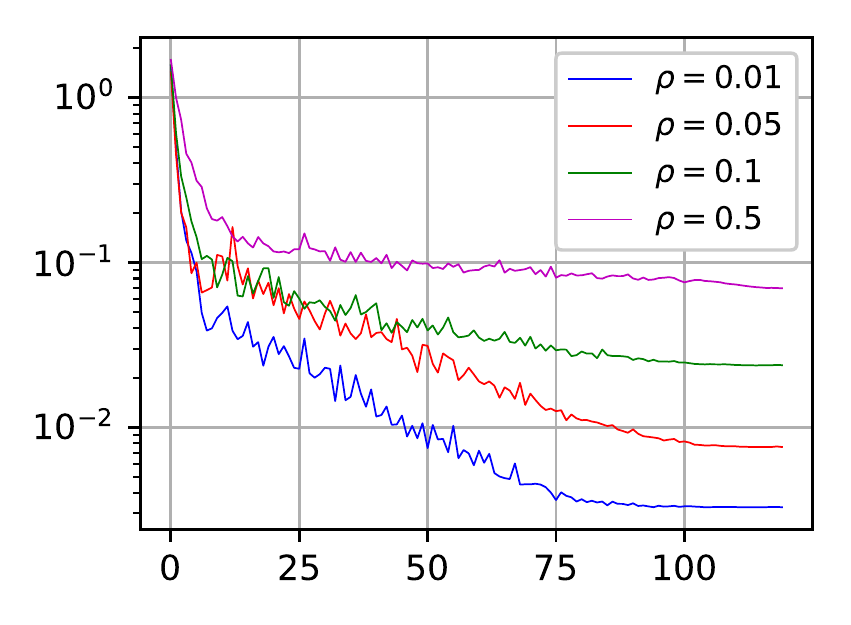}
	            \subcaption{\small $\ell(\cW)$}
	            \label{fig:conv:a}
	    \end{subfigure}
         \begin{subfigure}{0.63\columnwidth}
             \includegraphics[width=\linewidth]{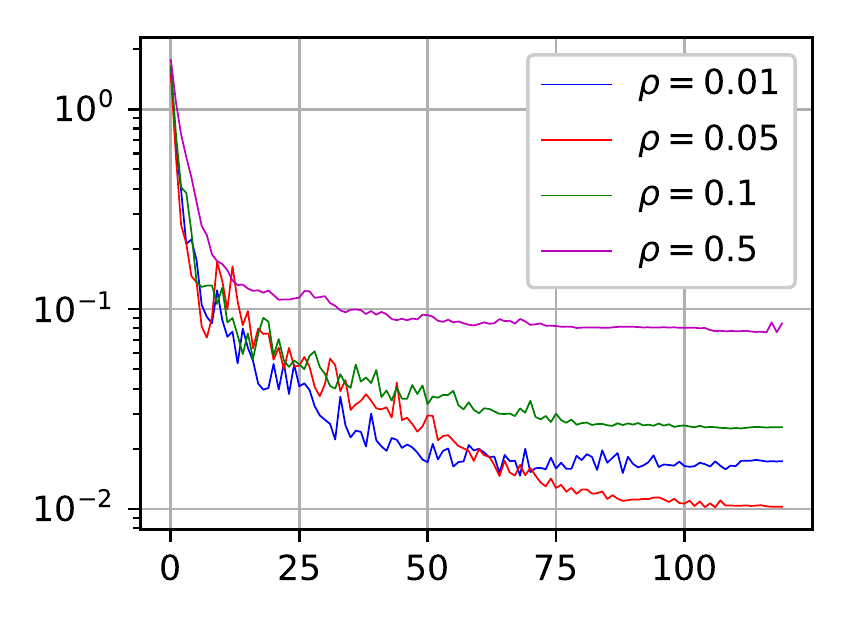}
                     \subcaption{\small $\ell(\textrm{Quantized}(\cW))$}
                     \label{fig:conv:b}
         \end{subfigure}
          \begin{subfigure}{0.63\columnwidth}
             \includegraphics[width=\linewidth]{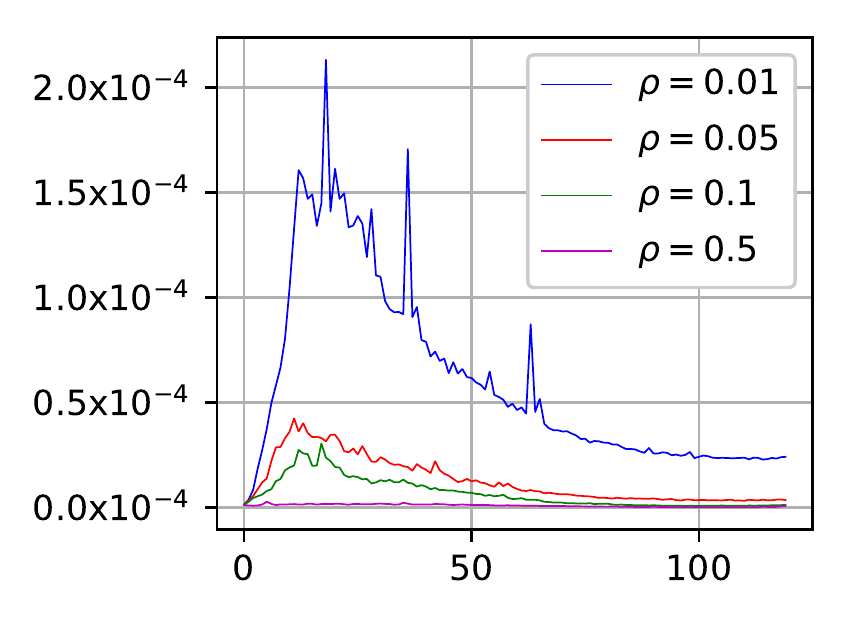}
          	\subcaption{\small $\text{MSE}(\cW, \cV)$.}
	\label{fig:conv:c}
         \end{subfigure}
         \caption{Training loss and MSE (between $\cW$ and $\cV$) with different values of $\rho$.}
	\label{fig:convergence}
\end{figure*}

\subsection{Convergence and Sensitivity of $\rho$}
To address the impact of hyper-parameter $\rho$ and the convergence of our training algorithm based on ADMM, we plot the training curves on MNIST classification experiments with various $\rho \in \{0.01, 0.05, 0.1, 0.5\}$ in Figure~\ref{fig:convergence}. Figure~\ref{fig:conv:a} shows the training loss of $\cW$, Figure~\ref{fig:conv:b} shows the training loss of the quantized $\cW$, where the bitwidth is set according to $\cV$. We can see that $\ell(\cW)$ converges to smaller values with smaller $\rho$, since smaller $\rho$ emphasizes more on the primal loss term. If perform the quantization on $\cW$, the smallest loss is not achieved by the smallest $\rho$ anymore, this is because $\cW$ is not well constrained with the quantized structure when $\rho$ is too small. To evaluate how the variables $\cW$ differs from $\cV$, we show the mean square error (MSE) between $\cW$ and $\cV$ in~Figure~\ref{fig:conv:c}. We can see that the MSE curves usually increase in the beginning and then decrease, and $\rho=0.05$ is enough to make the MSE $\rightarrow 0$.

\begin{table*}[htbp]
	\Huge
	\centering
	\caption{\small Comparison across different compression methods on ImageNet.}
	\renewcommand*{\arraystretch}{1.1}
	\renewcommand*{\tabcolsep}{10pt}
	\resizebox{1.9\columnwidth}{!}
	{
		\begin{tabular}{c|c|cccccccc}
			\toprule[0.15em]
			Model & Method & Automated & Pruning & Quantization & NZ\% & Ave. bits & Comp. rate& Acc.-1$\downarrow$ & Acc.-5$\downarrow$ \\
			\midrule[0.05em]
			MobileNet & Uniform Baseline~\citet{howard2017mobilenets} & \xmark & \cmark & \xmark &  61\% & - & 1.6$\times$ & 2.50\% & 1.70\% \\
			& Uniform Baseline~\citet{howard2017mobilenets} & \xmark & \cmark & \cmark &  61\% & 8 & 6.6$\times$ & 4.10\% & 2.90\% \\
			& Deep Compression~\citet{han2015deep} & \xmark & \xmark & \cmark &  - & 2 & 16$\times$ & 33.28\% & 25.59\% \\
			& HAQ~\citep{wang2019haq} & \cmark & \xmark & \cmark & - & 2 & 16$\times$ & 13.76\% & 8.03\% \\
			& {\bf Ours}  & \cmark & \xmark & \cmark & - & 2  & 16$\times$ & {7.10\%} & 4.40\%\\
			& Deep Compression~\citet{han2015deep} & \xmark & \xmark & \cmark & - & 3 & 10.7$\times$ & 4.97\% & 3.05\%\\
			& HAQ~\citep{wang2019haq} & \cmark & \xmark & \cmark & - & 3 & 10.7$\times$ & 3.24\% & 1.69\%\\
			& {\bf Ours} & \cmark & \xmark & \cmark & - & 3 & 10.7$\times$ & {\bf 1.19\%} & {\bf 0.76\%}\\
			 & {\bf Ours} & \cmark & \cmark & \cmark &  42\% & 2.8  & {\bf26.7}$\times$ & {4.41\%} & 2.61\%\\
			\hline
			AlexNet & Constraint-Aware~\citet{chen2018constraint} & \cmark & \cmark & \xmark & 4.9\% & - & 20$\times$ & 2.57\% & -\\
			& Deep Compression~\citet{han2015deep} & \xmark & \cmark & \cmark & 11\% & 5.4 & 54$\times$ & 0.00\% & -0.03\%\\
			 & CLIP-Q~\citep{tung2018clip} & \cmark & \cmark & \cmark & 8\% & 3.3 & 119$\times$ & -0.70\% & -\\
			 & {\bf Ours} & \cmark & \cmark & \cmark & 7.4\% & 3.7 & {118}$\times$ & {\bf -1.00\%} & {\bf -1.15\%}\\
			 & Ye et al.~\citet{ye2018unified} & \xmark & \cmark & \cmark & 4\% & 4.1 & {\bf 210$\times$} & 0.10\% & -\\
			 & {\bf Ours} & \cmark & \cmark & \cmark & 5\% & 3.1 & {205}$\times$ & {-0.08\%} & -0.56\%\\
			 \hline
			 MnasNet & Fixed-Bitwidth & \cmark & \cmark & \cmark & 50\% & 4 & 16$\times$ & {3.14\%} & {1.86\%}\\
			& {\bf Ours} & \cmark & \cmark & \cmark & 50\% & 3.7 & 17.1$\times$ & {\bf 1.66\%} & {\bf 0.92\%}\\
			& {\bf Ours} & \cmark & \cmark & \cmark & 30\% & 3.0 & {\bf 35.6$\times$} & 5.82\% & 3.23\%\\
			\hline
			ProxylessNAS- & Fixed-Bitwidth & \cmark & \cmark & \cmark & 50\% & 4 & 16$\times$ & {3.17\%} & {1.73\%}\\
			mobile & {\bf Ours} & \cmark & \cmark & \cmark & 51\% & 3.8 & 16.8$\times$ & {\bf 2.13\%} & {\bf 1.16\%}\\
			& {\bf Ours} & \cmark & \cmark & \cmark & 31\% & 2.9 & {\bf 35.6$\times$} & 5.21\% & 2.84\%\\
			\bottomrule[0.15em]
		\end{tabular}
	}
	\label{tab:imagenet}
	\vspace{-8pt}
\end{table*}

\subsection{Comparisons with State-of-the-arts}

\paragraph{ImageNet} In Table~\ref{tab:imagenet}, we show the validation accuracies of compressed models of different methods on ImageNet classification. We list the nonzero weights percentage, averaged bitwidth, the compression rate (original weights size / compressed weights size), and the (top-1 / top-5) accuracy drop. For MobileNet, we compare with the quantization methods of Deep Compression~\citet{han2015deep} and HAQ~\citep{wang2019haq}. We also compare with the uniform compression baselines~\citep{howard2017mobilenets}. The original MobileNet has $70.9\%$ top-1 accuracy and $89.9\%$ top-5 accuracy. Our quantization-only results with averaged bitwidth 2 and 3 have $7.1\%$ and $1.19\%$ top-1 accuracy drops respectively, which are about $2\times$ smaller than the HAQ counterparts ($13.76\%$ and $3.24\%$). The compression rate can be further improved to $26.7\times$ when jointly perform pruning and quantization.

For AlexNet, we compare with pruning or joint pruning and quantization methods. Unlike our end-to-end framework, all the compared methods set the pruning ratios and quantization bitwidth as hyper-parameters. Constraint-Aware Compression~\citep{chen2018constraint} and CLIP-Q~\citep{tung2018clip} uses Bayesian optimization to choose these hyper-parameters, while others manually set them. The uncompressed AlexNet is from PyTorch pretrained models and has $56.52\%$ top-1 accuracy and $79.07\%$ top-5 accuracy. When compressing the model to be 118$\times$ smaller, our method has an $1\%$ top-1 accuracy improvement which is higher than the compressed CLIP-Q model with similar compression rate. Our method can also compress AlexNet to be 205$\times$ smaller without accuracy drop, while the compressed model of Ye~et~al.~\citet{ye2018unified} has a $0.1\%$ top-1 accuracy drop with a similar compression rate.

For the NAS-based compact models, the uncompressed MnasNet has $73.46\%$ top-1 accuracy and $91.51\%$ top-5 accuracy, and the uncompressed ProxylessNAS-mobile has $74.59\%$ top-1 accuracy and $92.20\%$ top-5 accuracy.
We also evaluated a joint pruning and quantization baseline (Fixed-Bitwidth) by fixing the bitwidth for all the layers as 4 and pruning 50\% weights based on magnitude~\citep{han2015deep}.
Compared with AlexNet, we can find that the accuracies on these compact models are easier to be influenced by compression. This phenomenon is similar as in MobileNet.
%
%For AlexNet, we compare with methods of joint pruning and quantization. Unlike our end-to-end framework, all the compared methods set the pruning ratios and quantization bitwidth as hyper-parameters. CLIP-Q~\citep{tung2018clip} uses Bayesian optimization to choose these hyper-parameters, while others manually set them. The uncompressed AlexNet is from PyTorch pretrained models and has $56.52\%$ top-1 accuracy. When compressing the model to be 118$\times$ smaller, our method has an $1\%$ accuracy improvement which is higher than the compressed CLIP-Q model with similar compression rate. Our method can also compress AlexNet to be 205$\times$ smaller without accuracy drop, while the compressed model of Ye~et~al.~\citet{ye2018unified} has a $0.1\%$ accuracy drop with a slightly lower compression rate.

\paragraph{MNIST} Table~\ref{tab:mnist} shows the results of LeNet-5 on MNIST. The accuracy of the uncompressed LeNet-5 is $99.2\%$. Both Ye~et~al.~\citet{ye2018unified} and our method can achieve about $2000\times$ compression rate, while our compressed model does not have accuracy drop. Compare with the detail of its compressed model, we find that our method tends to leave more nonzero weights but uses less bits to represent each weight.
\begin{table}[htbp]
	\Huge
	\centering
	\caption{\small Comparison across different compression methods on LeNet-5@MNIST. All the methods adopt both pruning and quantization.}
	\renewcommand*{\arraystretch}{1.1}
	\renewcommand*{\tabcolsep}{6pt}
	\resizebox{\columnwidth}{!}
	{
		\begin{tabular}{c|cccccc}
			\toprule[0.15em]
			Method & Automated & NZ\% & Avg. bits & Comp. rate& Acc.$\downarrow$ \\
			\midrule[0.05em]
			Deep Compression~\citep{han2015deep} & \xmark & 8.3\% & 5.3 & 70$\times$ & 0.1\% \\
			BC-GNJ~\citep{louizos2017bayesian} & \xmark & 0.9\% & 5 & 573$\times$ & 0.1\%\\
			BC-GHS~\citep{louizos2017bayesian} & \xmark & 0.6\% & 5 & 771$\times$ & 0.1\%\\
			Ye et al.~\citet{ye2018unified} & \xmark & 0.6\% & 2.8 & 1,910$\times$ & 0.1\% \\
			{\bf Ours} & \cmark & 1.0\% & 1.46 & {\bf 2,120}$\times$ & {\bf0.0\%} \\
			\bottomrule[0.15em]
		\end{tabular}
	}
	\label{tab:mnist}
\end{table}

\begin{figure*}[tp]
         \begin{center}
         \begin{subfigure}[b]{\figwidthly\linewidth}
             \includegraphics[width=1\linewidth]{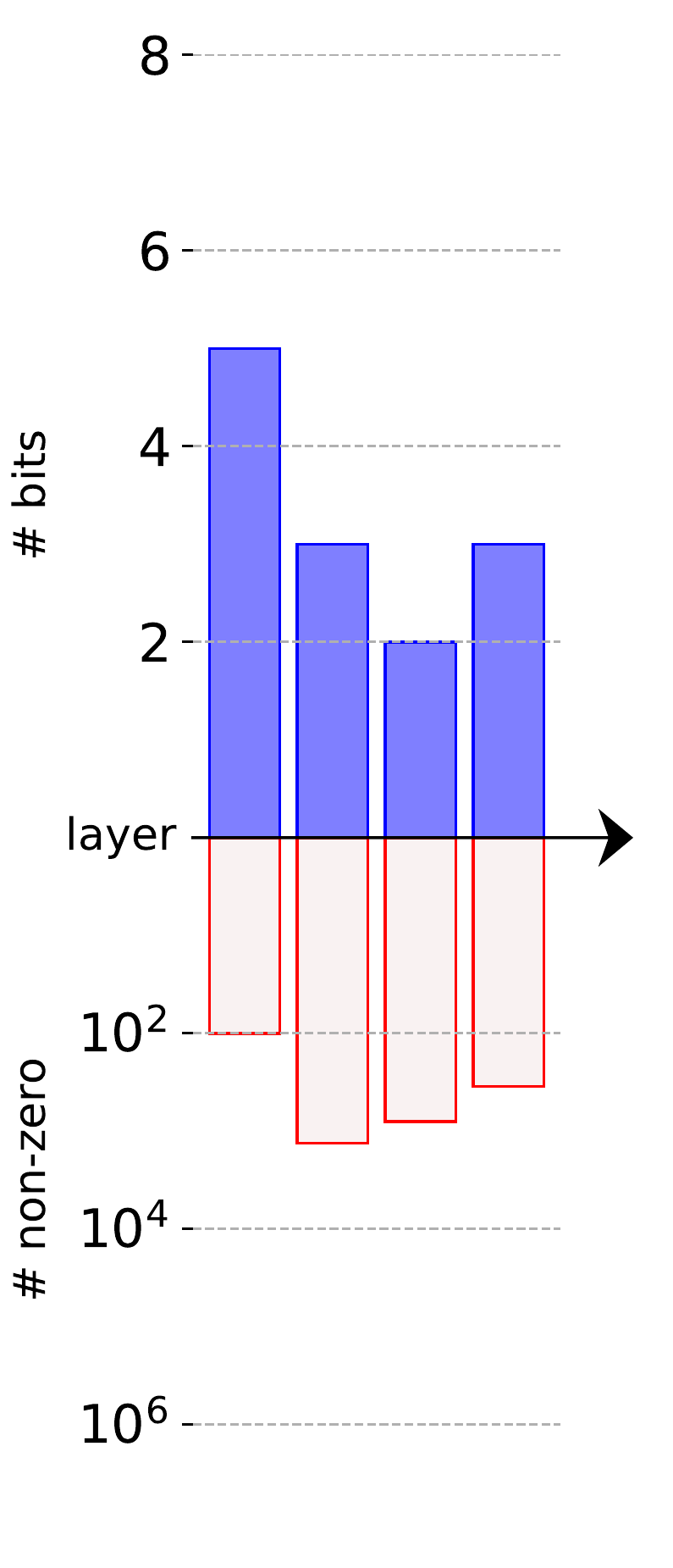}
                     \subcaption{\scriptsize LeNet-5~(\citet{ye2018unified})}
                     \label{fig:vis:a}
         \end{subfigure}
         \begin{subfigure}[b]{\figwidthlo\linewidth}
           \includegraphics[width=1\linewidth]{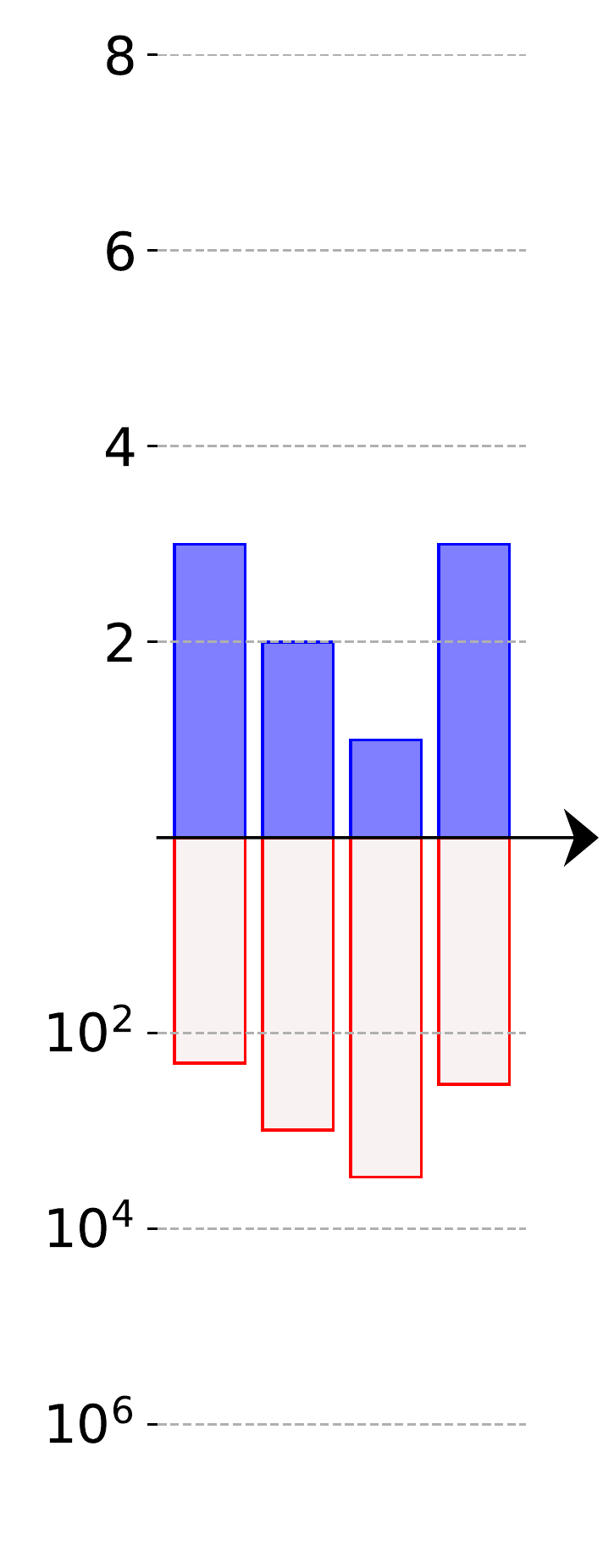}
                   \subcaption{\scriptsize LeNet-5~(Ours)}
                   \label{fig:vis:b}
         \end{subfigure}
          \begin{subfigure}[b]{\figwidtha\linewidth}
           \includegraphics[width=1\linewidth]{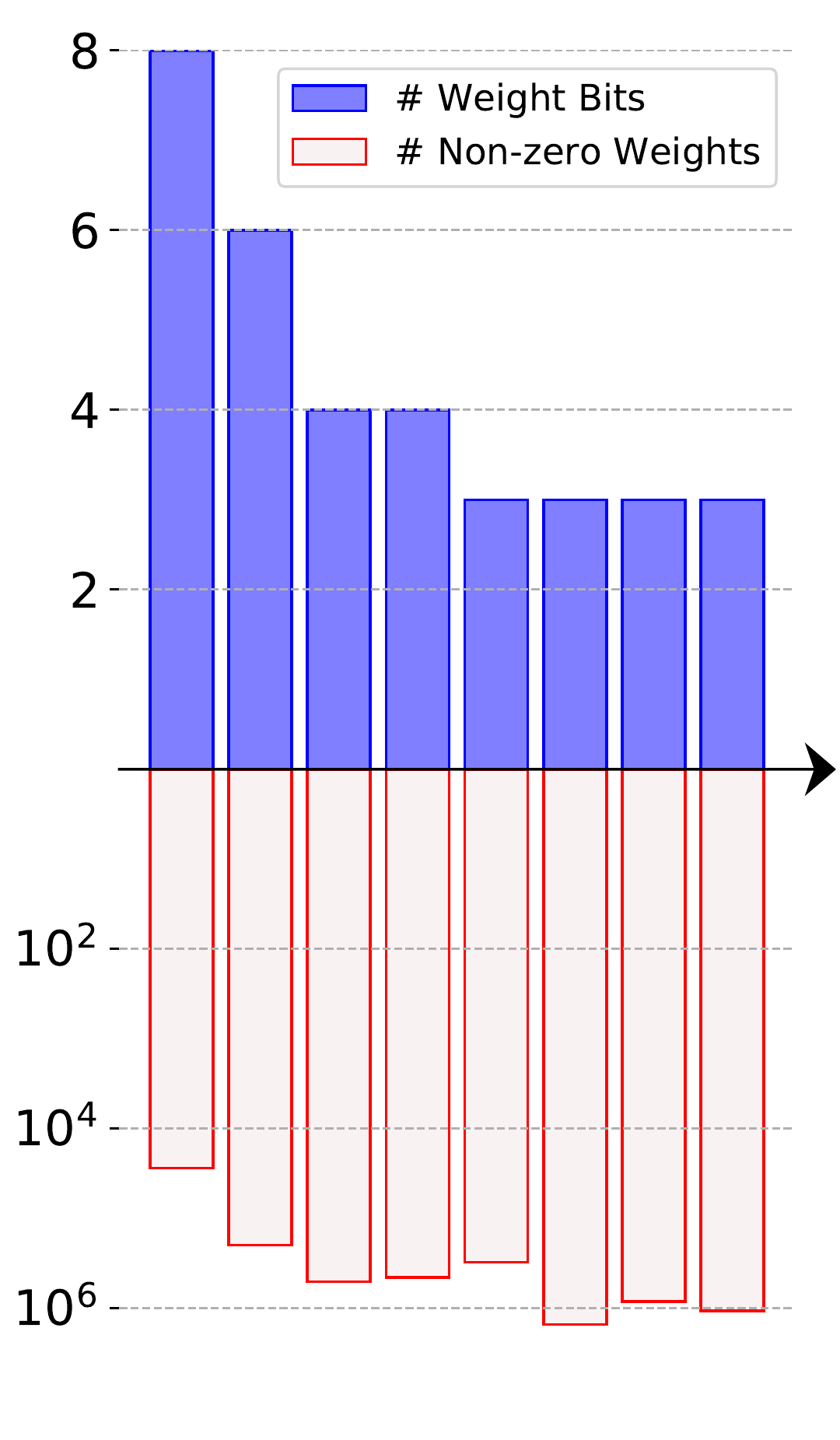}
                   \subcaption{\scriptsize AlexNet (119$\times$\citep{tung2018clip})}
                   \label{fig:vis:c}
         \end{subfigure}
         \begin{subfigure}[b]{\figwidtha\linewidth}
           \includegraphics[width=1\linewidth]{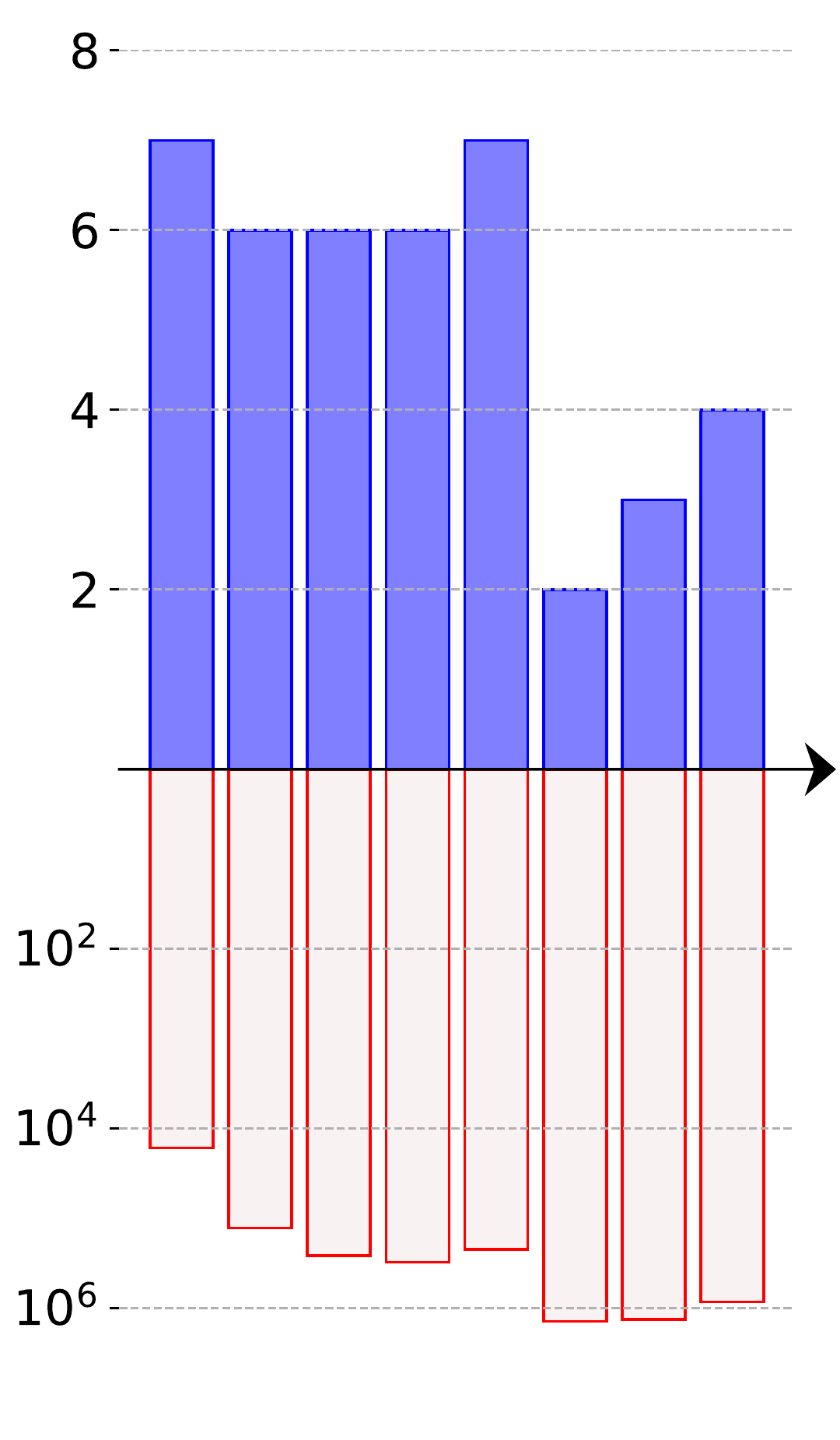}
                   \subcaption{\scriptsize AlexNet~(Ours 118$\times$)}
                   \label{fig:vis:d}
         \end{subfigure}
         \begin{subfigure}[b]{\figwidtha\linewidth}
             \includegraphics[width=1\linewidth]{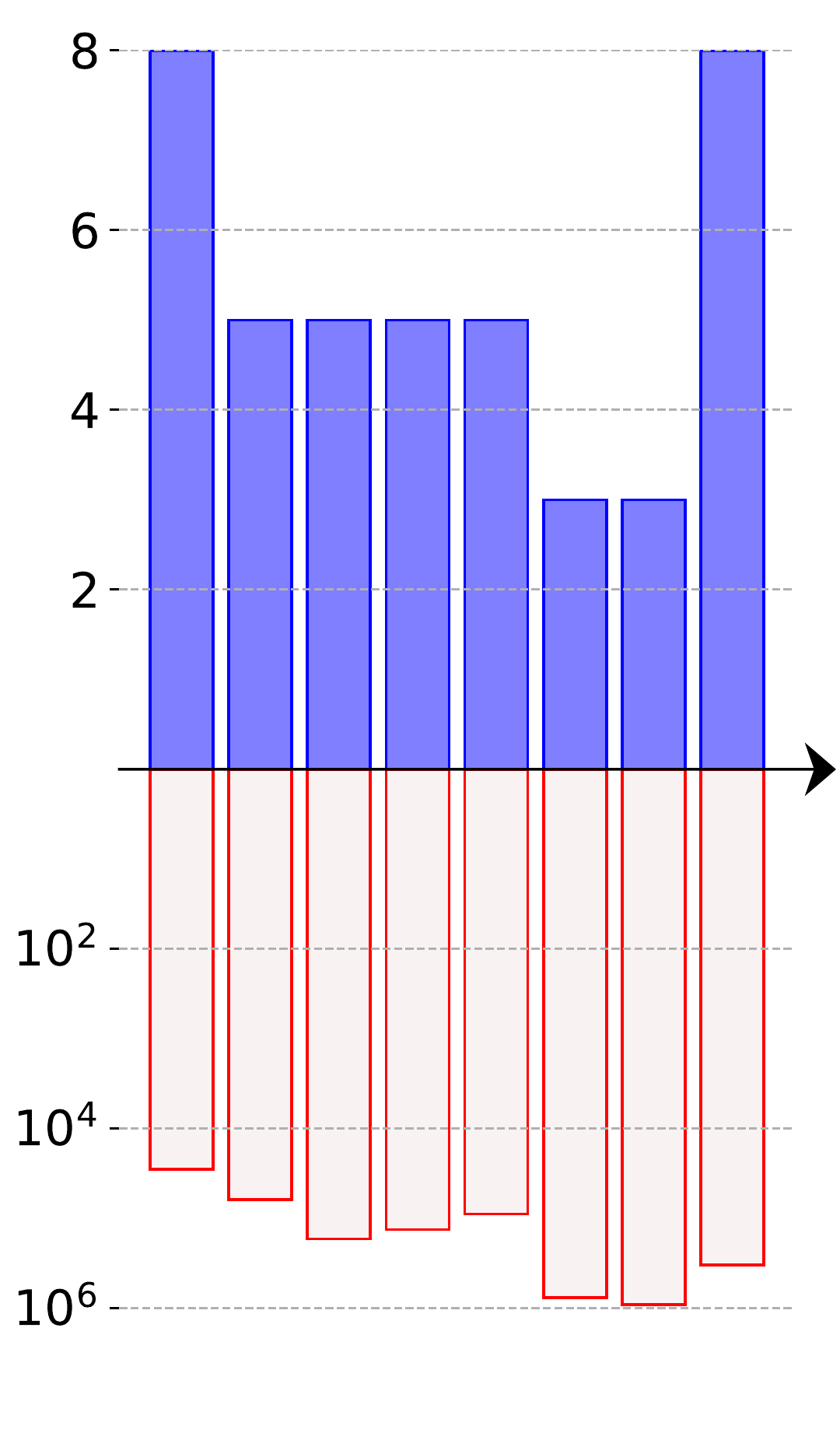}
                     \subcaption{\scriptsize AlexNet~(210$\times$\citet{ye2018unified})}
                     \label{fig:vis:e}
         \end{subfigure}
         \begin{subfigure}[b]{\figwidtha\linewidth}
           \includegraphics[width=1\linewidth]{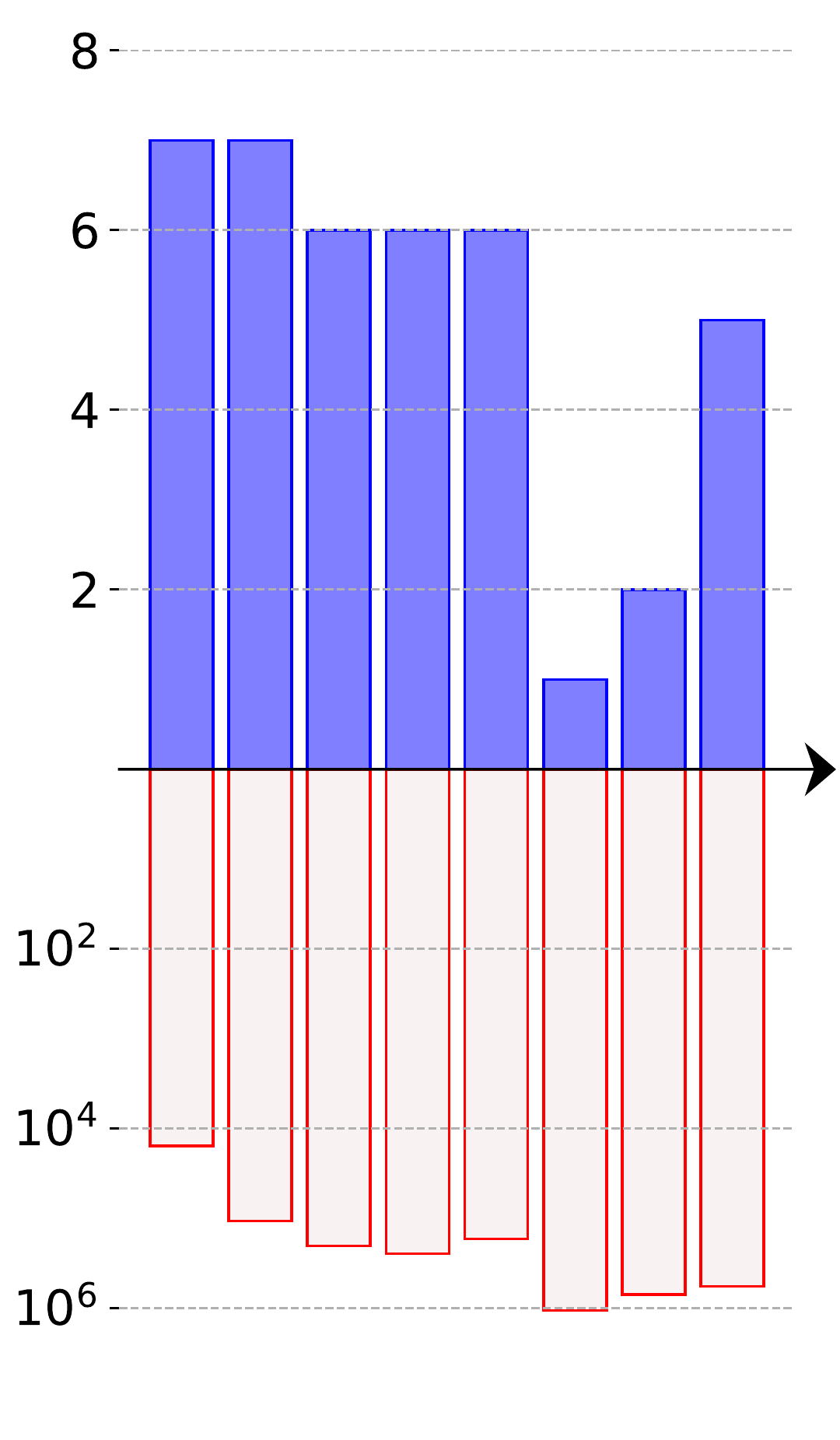}
                   \subcaption{\scriptsize AlexNet~(Ours 205$\times$)}
                   \label{fig:vis:f}
         \end{subfigure}
         \caption{Visualization of the compressed results of different layers on LeNet-5 and AlexNet. The number of nonzero weights is shown in $\log_{10}$ scale. Our compressed models are presented in (b), (d) and (f) to compare with the network compressed by CLIP-Q~\cite{tung2018clip} and Ye~et~al.~\cite{ye2018unified}.
         }
         \label{fig:vis}
         \end{center}
         \vspace{-15pt}
 \end{figure*}

\paragraph{CIFAR-10} Table~\ref{tab:cifar} shows the results of the compressed ResNets on CIFAR-10 dataset. The accuracy of the original ResNet-20 is $91.29\%$ and the accuracy of ResNet-50 is $93.55\%$. For ResNet-20, we compare with the automated quantization method ReLeQ~\citep{yazdanbakhsh2018releq}. For fair comparison, we evaluate two compressed models of our method, one only uses quantization and another uses jointly pruning and quantization. For the quantization-only model, we achieve $16\times$ compression rate without accuracy drop, which has better accuracy and smaller size than ReLeQ. When introducing pruning, there is a $0.14\%$ accuracy drop but the compression rate is improved to $35.4\times$.

\begin{table}[hbp]
	\Huge
	\centering
	\caption{\small Comparison across different methods on CIFAR-10. All the methods automatically set the compression ratios.}
	\renewcommand*{\arraystretch}{1.1}
	\renewcommand*{\tabcolsep}{10pt}
	\resizebox{\columnwidth}{!}
	{
		\begin{tabular}{c|c|cccccc}
			\toprule[0.15em]
			Model & Method & Pruning & Quantization & NZ\%  & Ave. bits & Comp. rate& Acc.$\downarrow$ \\
			\midrule[0.05em]
			ResNet-20 & ReLeQ~\citep{yazdanbakhsh2018releq} & \xmark &\cmark & - & 2.8 & 11.4$\times$ & 0.12\% \\
			 & {\bf Ours} & \xmark &\cmark & - & 2 & 16$\times$ & {\bf 0.00}\%\\
			 & {\bf Ours} & \cmark &\cmark & 46\% & 1.9  & {\bf35.4}$\times$ & {0.14\%}\\
			\hline
			ResNet-50 & AMC~\citep{he2018amc} & \cmark &\xmark & 60\% & - & 1.7$\times$ & -0.11\% \\
			 & {\bf Ours} & \cmark &\xmark & 50\% & - & 2$\times$ & {\bf-1.51}\%\\
			 & {\bf Ours} & \cmark &\cmark & 4.2\% & 1.7 & {462}$\times$ & {-1.25\%} \\
			 & {\bf Ours} & \cmark &\cmark & 3.1\% & 1.9 & {565}$\times$ & {-0.90\%} \\
			 & {\bf Ours} & \cmark &\cmark & 2.2\% & 1.8 & {\bf836}$\times$ & {0.00\%} \\
			\bottomrule[0.15em]
		\end{tabular}
	}
	\label{tab:cifar}
	\vspace{-5pt}
\end{table}

For ResNet-50, we compare with the automated pruning method AMC~\citep{he2018amc}. Its compressed ResNet-50 targeted on model size reduction has $60\%$ of non-zero weights. In our experiment, we find that ResNet-50 still has a large space to compress. The pruning-only result of our method compress ResNet-50 with $50\%$ weights and an $1.51\%$ accuracy improvement. By performing jointly pruning and quantization, our method can compress the ResNet-50 with compression rate from $462\times$ to $836\times$. The accuracy loss is only met when compress the model to $836\times$ smaller, which suggests the ResNet-50 is mostly redundant on CIFAR-10 classification, and compressing it could reduce overfitting.

\paragraph{Compressed model visualization} In Figure~\ref{fig:vis}, we visualize the distribution of sparsity and bitwidth for each layer on LeNet-5 and AlexNet. Subfigures~\ref{fig:vis:a}, \ref{fig:vis:c} and \ref{fig:vis:e} show compressed models of Ye~et~al.~\citet{ye2018unified} and CLIP-Q~\citep{tung2018clip}. Subfigures~\ref{fig:vis:b}, \ref{fig:vis:d} and \ref{fig:vis:f} are our compressed models. For LeNet-5, we observe that our method preserves more nonzero weights in the third layer, while allocates less bitwidth compared with Ye~et~al.~\citet{ye2018unified}. For AlexNet, our method has the trend of allocating larger bitwidth to convolutional layers than fully connected layers. CLIP-Q also allocates more bits to the convolutional layers, while Ye~et~al.~\citet{ye2018unified} assign more bits to the first and last layer. Our method also shows a preference for allocating more bits to sparser layers. This coincides with the intuition that the weights of sparser layers may be more informative, and increasing the bitwidth on these layers also brings less storage growth.

\section{Conclusion}
As DNNs are increasing deployed on mobile devices, model compression is becoming more and more important in practice. Although many model compression techniques have been proposed in the past few years, lack of systematic approach to automatically set the layer-wise compression ratio diminishes their performance. Traditional methods require human labor to manually tune the compression ratios. Recent work uses black-box optimization to search the compression ratios but introduces instability of black-box optimization and is not efficient enough. 
%Different from prior work, we try to start from the root of the problem. Specifically, 
We propose a constrained optimization formulation which considers both pruning and quantization and does not require compression ratio as hyper-parameter.
By using ADMM, we build a framework to solve the constrained optimization problem efficiently. Experiment shows our method outperforms the handcrafted and hyper-parameter search approaches.
%Efficient algorithms for Knapsack problems are introduced to solve the sub-procedures in ADMM.
%In the future, we want to investigate other scenarios where we can substitute black-box optimization for more efficient optimization based approaches.

\section*{Acknowledgement}
We gratefully acknowledge supports from NSF CCF Award \#1714136.

{\small
\bibliographystyle{ieee_fullname}
\bibliography{reference}
}
\end{document}